\documentclass{article}
\usepackage[preprint]{neurips_2019}
\usepackage{graphicx}
\usepackage[utf8]{inputenc} 
\usepackage[T1]{fontenc}    
\usepackage{hyperref}       
\usepackage{url}            
\usepackage{booktabs}       
\usepackage{amsfonts}       
\usepackage{nicefrac}       
\usepackage{microtype}      
\usepackage{wrapfig}
\usepackage{color}
\usepackage{xcolor}
\usepackage{hyperref}

\usepackage{authblk}

\title{ \textbf{Mask Embedding in conditional GAN for Guided Synthesis
of High Resolution Images} }

\author[1,2]{Yinhao Ren}
\author[2]{Zhe Zhu}
\author[3]{Yingzhou Li}
\author[4]{Joseph Lo}

\affil[1]{Department of Biomedical Engineering, Duke University}
\affil[1,2,4]{Department of Radiology, Duke University School of Medicine}
\affil[3]{Department of Mathematics, Duke University}
\affil[]{\textit {\{yinhao.ren, zhe.zhu, yinzhou.li, joseph.lo\}@duke.edu}}

\begin{document}

\maketitle

\begin{abstract}
Recent advancements in conditional Generative Adversarial Networks
(cGANs) have shown promises in label guided image synthesis. Semantic
masks, such as sketches and label maps, are another intuitive and
effective form of guidance in image synthesis. Directly incorporating
the semantic masks as constraints dramatically reduces the variability
and quality of the synthesized results. We observe this is caused by
the incompatibility of features  from different  inputs (such as mask
image and latent vector) of the generator. To use semantic masks as
guidance whilst providing realistic synthesized results with fine
details, we propose to use mask embedding mechanism  to allow for
a more efficient  initial feature projection  in the generator. We
validate the effectiveness of our approach by training a mask guided
face generator using  CELEBA-HQ dataset. We can generate realistic and
high resolution facial images up to the resolution of $512\times 512$
with a mask guidance. Our code is publicly available\footnote{\url{https://github.com/johnryh/Face_Embedding_GAN}}.
\end{abstract}

\section{Introduction} \label{Introduction}
The ability to synthesize photo-realistic images from a semantic
map is highly desired for various image editing applications. Most
existing approaches with semantic mask inputs focus on either
applying the coarse to fine synthesis with a cascade of networks
\cite{Pix2pix_HD, refinementGAN, stackGAN}, or designing specific
loss functions \cite{wGAN, wgan_gp} to  increase the model stability
for better image quality. Though advances have been made, it is still
challenging to synthesize  high resolution images with diverse local features using semantic masks as guidance.

In this work, we propose a novel technique that enables the
generative models to synthesize images that are coherent with the
provided semantic mask constraint while preserving the diversity of
local texture details. This characteristic is especially useful in
image generation applications that require high resolutions output,
feature diversity and high fidelity. For example, a live art editing
interface implemented with this technique would allow content creators
to focus  on the global concept while the algorithm deals with local
details automatically.

\begin{figure}
  \centering
  \label{fig:One_to_More_Mapping}
\end{figure}

\begin{figure}[!ht]
    \setlength\tabcolsep{0.25pt}
    \includegraphics[width=0.33\textwidth]{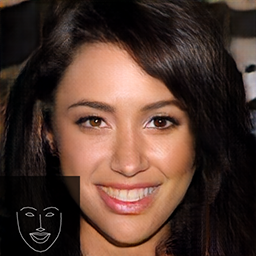}
    \includegraphics[width=0.33\textwidth]{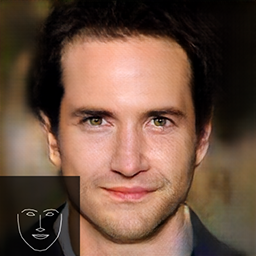}
    \includegraphics[width=0.33\textwidth]{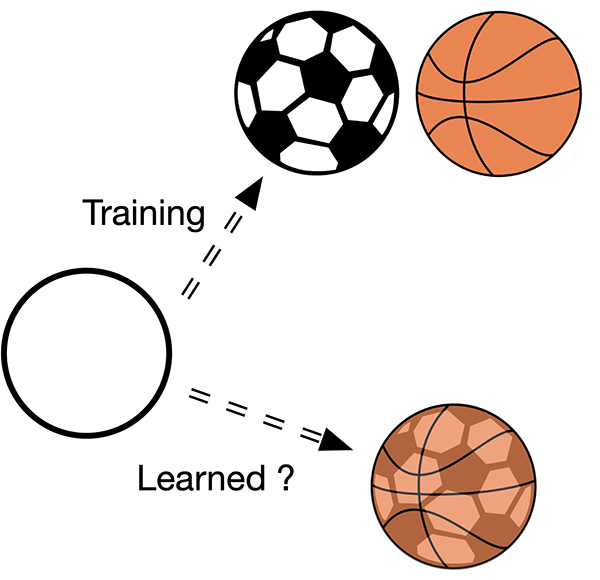}
    \label{fig:teaser}
    \caption{Generated image samples and cartoon illustration of the sample space mapping challenge during training a mask guided face generator: An image translation model trained to map the same
    circle pattern to various different ball patterns. The model
    is likely to learn the "average" ball pattern of the training
    dataset instead of been able to map to all of them individually. The same problem exist for training a generator to replicate different faces with very similar mask representation.}
\end{figure}

Image translation models such as Pix2Pix \cite{pix2pix2016}
that directly map the abstract representation of images with the
original images using U-Net \cite{UNet} style generator do not have
the proper mechanism for stochastic feature realizations. This
usually causes the model to output ambiguous features that are
in between possible solutions in the feature space, as shown in
Fig \ref{fig:One_to_More_Mapping}. Heuristically, this leads to
blurred images and ill-defined texture details. One solution
of this problem is proposed by Wang et al.~\cite{Pix2pix_HD}
(Pix2Pix-HD). They  use a coarse to fine approach together with
perceptual loss to refine the output image quality. However, this
approach requires a much larger model, but still does not solve the
fundamental issue of feature mapping. A more theoretically sound
solution proposed by Zhao et al.~\cite{Tube_GAN} (Tube-GAN)  is to
use both a latent vector $z$ (noise in a chosen distribution) and a
semantic mask as conditional input, allowing the model to learn the
joint distribution. Nevertheless, their proposed merging strategy
of projected latent features and projected mask features are not
by default guaranteed to be coherent, thus this model is limited to
generating only low frequency information in the background.
Thus it is still a challenging task to use pixel-level mask as guidance
to generate high resolution images with fine details.

In this paper, we address the two main issues of current state-of-art
mask guided generative models~\cite{Pix2pix_HD,Tube_GAN} with
pixel-level semantic input: (1) lack of diverse fine-grained texture
details in synthesized results, caused by inefficient mask-to-image
domain mapping and (2) low parameter efficiency in current multiple
conditional inputs architecture designs. For the first issue, we
argue that coupling latent vector with the input semantic map leads to
better sample space mapping, which results in diverse texture details
in synthesized results. For the second issue, our solution is to inject
mask embedding into the latent input vector before the initial feature
projection, and this operation significantly improves the quality of
texture details in synthesized results. Coupling the mask embedding
vector with latent vector is an efficient way to add mask constrains,
since it allows the initial feature projection to be compatible with
the pixel-level mask constraint. Contrary to Tube-GAN we use the
projected mask features mainly as a constraint to the latent features
so that the up-sampling path of the network is able to preserve most
of its capacity to perform refinement of local texture details. Fig \ref{fig:teaser} provides a preview of our synthesized images and the corresponding input masks. The methods
and detailed analysis will be presented below. 

\section{Related Work}

\subsection{Conditional GAN}

Conditional GANs \cite{ConditionalGA} achieve the control of
generator output through coupling of latent vector and conditional
inputs. Many studies \cite{cGAN_face_attribute, cGAN_facial_expression,
cGAN_face_aging, cGAN_histology, cGAN_brain_tumor, cGAN_MRI}
applied cGAN using image attributes in vector form (such as labels)
for controlled image synthesis. Pix2Pix~\cite{pix2pix2016} and
Pix2Pix-HD~\cite{Pix2pix_HD} first proposed to use semantic input
directly in an encoder-decoder style structure for image-to-image
translation. Some studies have applied input embedding to transform a
higher dimensional attribute such as semantic mask into a more compact
lower dimensional form. CCGAN~\cite{text_embedding} proposes using
sentence embedding that contains image characteristics as feature
for cycle GAN training. Their study shows condensed text information
can be merged with generator latent features as conditions for image
synthesis. Yildirim et al.\cite{disentanglement} uses the binary
mask embedding as part of the conditional input to control the shape
of generated garment images. However, their work indicates mask
embedding vector is not sufficient for pixel-level mask constrained
image synthesis. The output shape of their proposed model does not
always align with the input mask.

\subsection{State-of-the-art Pix2Pix Style Generator}

\begin{figure}
  \centering
  \includegraphics[width=0.85\textwidth]{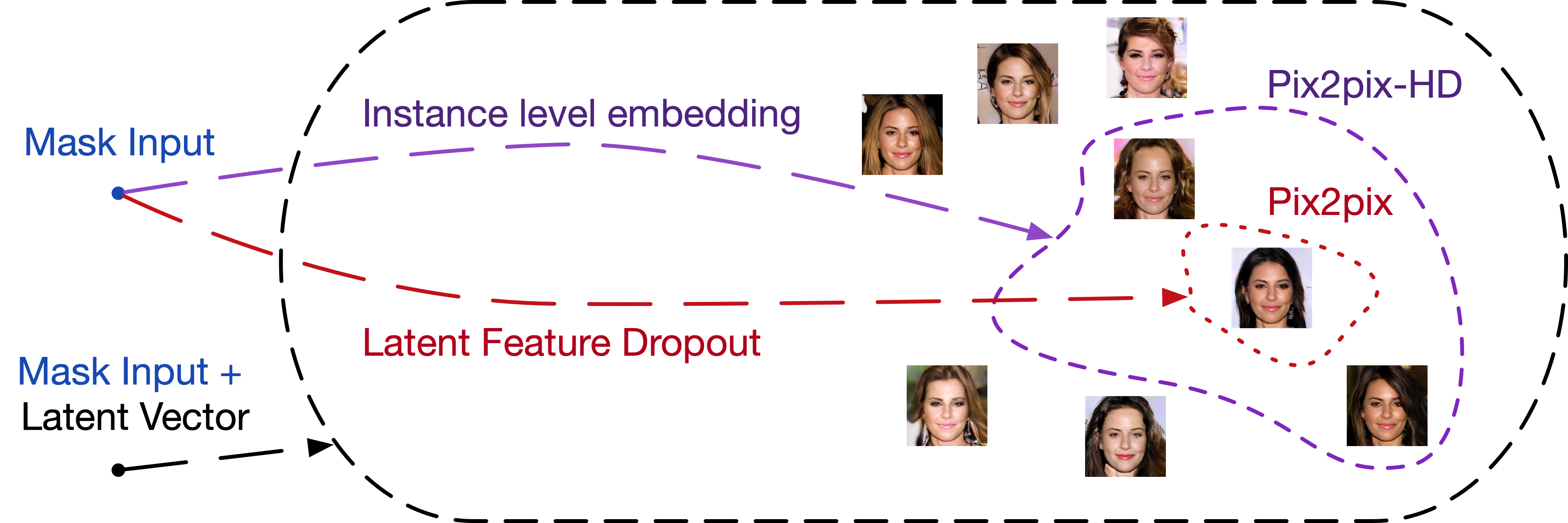}
  \caption{Reachable sample space under different mapping mechanisms.}
  \label{fig:Reachable_sample_space}
\end{figure}
Many works
\cite{pix2pix2016,Pix2pix_HD,Thermal_pix2pix,text_embedding,Retinal_GAN}
have applied image translation models to map the image from one domain
to another. However, a typical Pix2Pix setup does not perform well in
terms of fine-grained texture details and feature diversity. The state
of art Pix2Pix-HD model on the other hand proposes a coarse-to-fine
model architecture design with perceptual loss and multi-scale
discriminators. The main idea is to use additional loss terms to
regularize the expanded model capacity, especially the concatenated
refinement networks. Though the proposed model has a mechanism of
randomizing the textures through instance level feature embedding,
diversity of local texture details still relies on the minor
perturbations of instance label maps. To some extent this mechanism
allows stochastic texture generation, however, the mapping of textures
is coupled only with the shape perturbation of objects. In other words,
the image translation model is still limited to one-to-one  mapping
as shown in Fig \ref{fig:One_to_More_Mapping}, and the image quality
and diversity is rather low due to this limitation.

\subsection{Progressive Growing of GAN}

Progressive growing of GAN(pGAN)~\cite{karras2018progressive}
is a training methodology that gradually adds convolution layers
to the generator and discriminator to achieve better stability
and faster convergence. This technique makes it possible to
synthesize high resolution images using a slightly modified
DCGAN~\cite{DCGAN} generator and discriminator architecture. Several
recent studies~\cite{style_gan, pGAN_Brain, pgan_mammo} have applied
the progressive training strategy and achieved  high resolution of
synthesized results in non-conditional settings. We also apply progressive training strategy
to achieve high resolution outputs. 

\section{Mask Embedding in Generator}

To control the shape of generator output, a mask is typically used
as the only input in an encoder-decoder style generator to enforce
the pixel-level constraint. The fundamental principle of such image
translation models is to build a translation of $G(v)\rightarrow \{r\}$
where one to one translations are established give input $v$. With
mechanism such as drop-out or noise overlaid denoted as $z$ to input $v$, the one
to more relation $G(v,z) \rightarrow \{r_{1},r_{2} ... r_{m}\}$ becomes theoretically
possible in ideal cases. However, limited by the convolution operations
and choice of objective function, Pix2Pix reported that overlaid
noise is often ignored by the model. Model output typically depends
heavily on the semantic input mask and drop-out so that the diversity
of high frequency texture patterns is limited. In other words, given
a sub optimal image-to-image translation network $G^{'}$ the sampling
scheme in practice  becomes $G^{'}(v,z) \rightarrow \{r_{1},r_{2}
... r_{n}\}$ where $n \ll m$. The mapped sample space thus becomes
sparse. As illustrated in Fig \ref{fig:Reachable_sample_space}, we
postulate that these different strategies (Pix2Pix, Pix2Pix-HD, and
our proposed method) allow the model to sample increasingly larger
subsets of the entire domain space, which in turns provides greater
generator performance in terms of diversity, resolution and realism.

In this study we propose a new generator architecture design concept
that maps a particular semantic input to the sample space more
efficiently by coupling the latent input and the conditional input.

\subsection{Pixel-Level Mask Constraint and Model Design}

\begin{figure}
  \centering
  \includegraphics[width=1\textwidth]{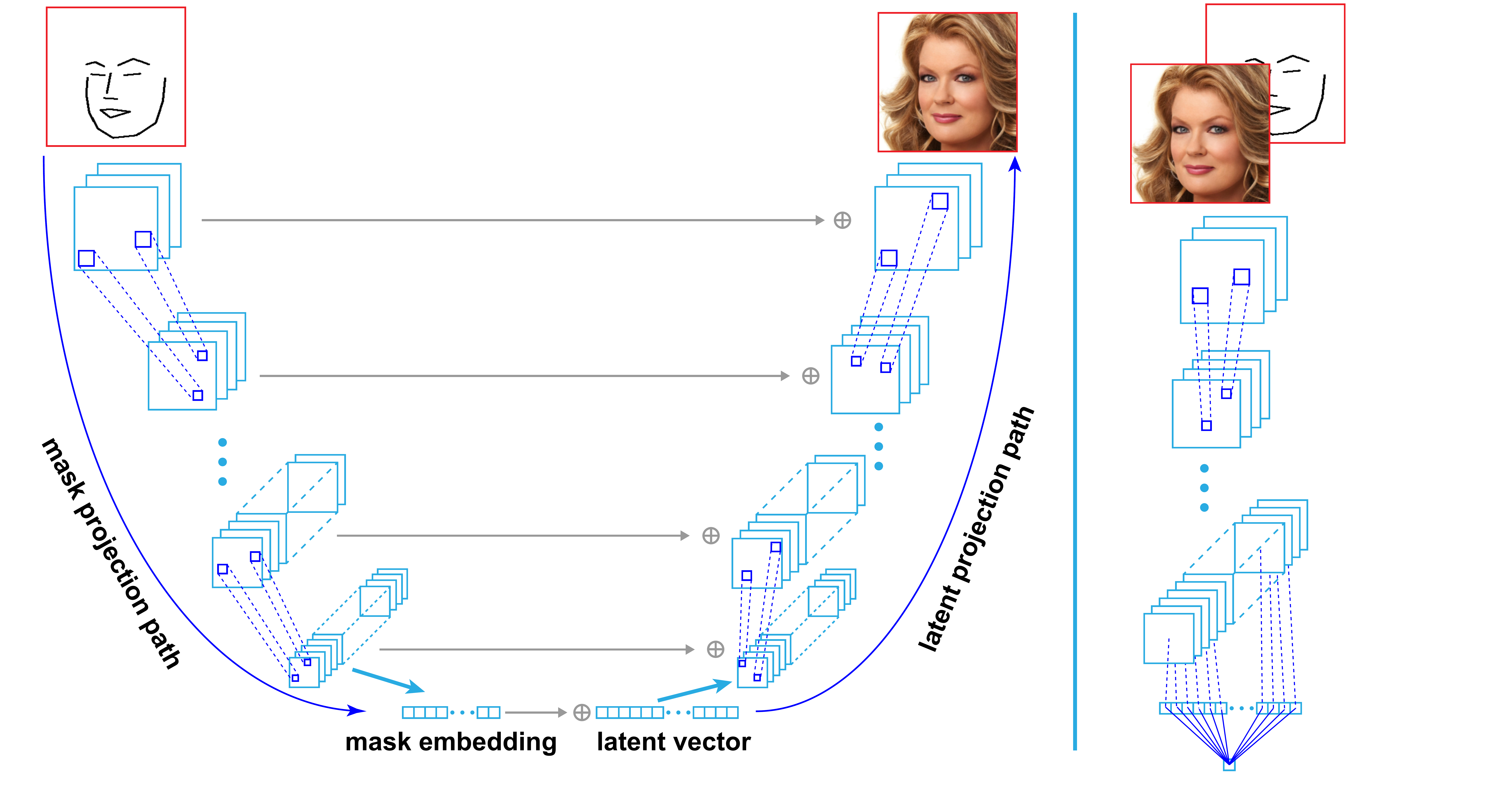}
  \caption{Architecture of our network. Left: a U-Net style
  generator. Right: a discriminator consists of several convolution
  layers.} \label{fig:architecture}
\end{figure}

Our proposed generator structure shown in Fig \ref{fig:architecture}
is derived from the pGAN generator architecture
\cite{karras2018progressive,DCGAN}, where the generator projects a
latent vector onto the latent space following several up-sampling
and convolution layers to form the output image. To inject the
semantic information, we construct a series of mask features and
concatenate them onto the corresponding latent features. This forms
the U-Net style architecture that is similar to the one implemented
in the Pix2Pix study but without the latent vector input. However, we
observe this initial implementation output images with significantly
reduced quality compared to the original pGAN framework. We regard
this issue as a space sampling problem where the mask is posing a
constraint on the feature projection path. The mapping of feature
values from early layers to later layers becomes less reliable with
the spatial and morphological constraints posed by the mask input,
resulting in reduced model capacity and unstable training process. A
reasonable solution is to implement a mechanism that allows the
initial latent feature projections to be mostly coherent with the mask
constraint. Then the model can use the short connections (horizontal
arrows in Fig~\ref{fig:architecture}) of mask features only as a
means to enforce the pixel-level constraint without consuming too
much model capacity to refine global image structures. We achieve
this mechanism by constructing a \emph{mask embedding} vector and injecting
it into the latent input vector, as shown in the bottom left of
Fig~\ref{fig:architecture}.

\subsection{Formulation} \label{Formulation}

\begin{figure}
  \centering
  \includegraphics[width=0.9\textwidth]{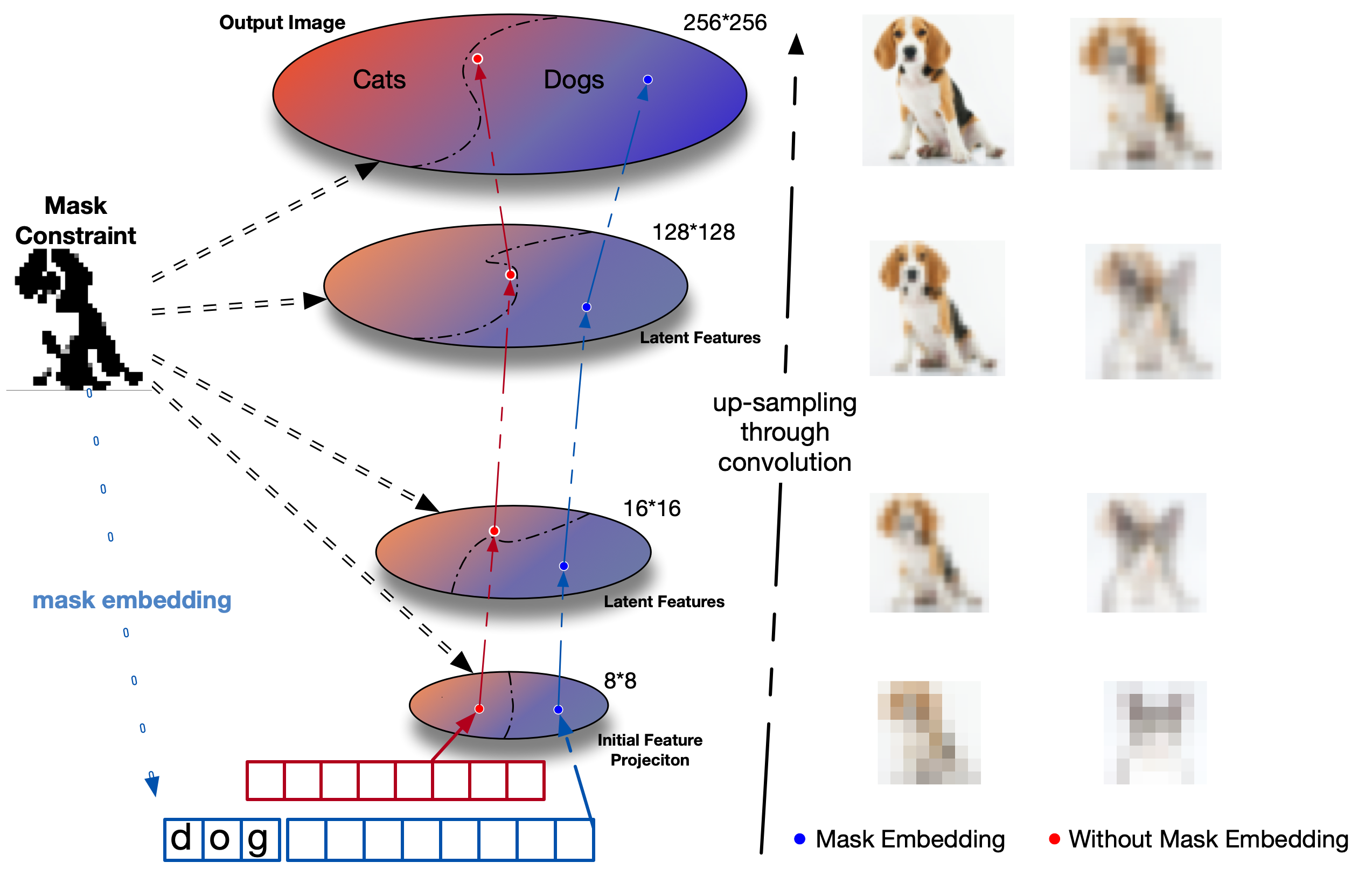}
  \caption{An illustrative example of generating an image of a dog
  using a dog mask as the guidance. Left: illustrative feature space of the
  image generation process using a series of convolution layers. Right:
  two examples of generating a dog image given a dog mask
  with and without mask embedding. For simplicity the latent features
  are visualized as low resolution images. At inference time,
  \textbf{an ideal model with mask embedding} projects base features
  onto the correct manifold and performs proper up-sampling through
  convolution layers; However, \textbf{model without mask embedding}
  learns to (1) project only average base image; (2) inefficiently map
  the average base image to a dog to comply with the mask constraint.}
  \label{fig:Sample_Space_convolution}
\end{figure}


This section aims to give some intuition behind our structure.
We regard having both mask constraint and local fine-grained
texture details at the same time as a space sampling problem, under the condition that
the up-sampling is mostly conducted with convolution layers.
A mask input does not identify an image in the real image dataset,
but instead relate to a cluster of real images.
Hence masks gathered from dataset defines a partition of the real image manifolds.
The top left ellipse in Fig~\ref{fig:Sample_Space_convolution} illustrates a two-partition of dog and cat masks.
We also demonstrates the low resolution feature sets by smaller ellipses.
Connected by a series of convolution layers, which is a local operation with limited receptive field,
the partition is inherited hierarchically and admits similar geometry within each manifold.
Our structure then aims to first correctly sample a mask constraint point in the lowest resolution manifold
in two steps, i.e., (1) locate the correct partition via mask embedding and (2) sample a point
within the partition via a latent feature vector.
Then an up-sampling procedure refines the detail and enhance the mask constraint
through the vertical injection of mask information as in Fig~\ref{fig:architecture}.
Two components, latent feature vector and mask embedding, are the fundamental difference
between our model and others. We would like to emphasize the importance of each of them.

Without latent feature vector, as in Pix2Pix or Pix2Pix-HD, the model with only mask input
does not have sufficient randomness. Hence the generated image is nearly uniquely identified by the mask.
The variety of the generated images through models without latent feature vector is very limited.
In contrast, our model have latent feature vector, which encodes a large variety of details.
Given a mask, we are able to generate dramatically different images still with fine details.

Without mask embedding, as in Tube-GAN, the constraint is less emphasized
in the lower dimensional features and parameters in the later layers potentially have to
correct incorrect low resolution image, which limits the capability in expressing details.
While our model uses mask embedding which potentially finds the correct partition
and generates correct latent space representation. Hence all later layers focus more on generating details.
Fig~\ref{fig:Sample_Space_convolution} shows a cartoon of the comparison. 
The blue dash line indicates the process of our model and generates dog image at different resolutions
in the second to the right column.
Whereas the red dash line indicates the process without mask embedding.
It generates a low resolution cat image in the beginning due to the lack of mask information.
During the later layers, convolution together with mask injection correct the image from cat to dog.
Unfortunately, the final image looks like a dog but is of much lower quality.
Images in Fig~\ref{fig:Sample_Space_convolution} are not generated by models,
but we do observe similar behavior in reality. These
observations indicate that incorporating the mask embedding
significantly improves the features projection efficiency.

\subsection{Architecture}

Our proposed model shown in Fig \ref{fig:architecture} consists
of the mask projection path and the latent projection path
corresponding to the contracting and expanding path in U-Net
\cite{UNet} respectively. The input to the mask projection path is
a binary face edge map. The mask undergoes a series of blocks, and
each block consists of 2 convolution layers with strides of 1 and 2
respectively. Each block outputs an increasing number of features to
the following layer and concatenate only the first 8 features to the
latent projection path to form the mask constraint.

The mask projection path has two main functions. First it provides
spatial constraint for the feature up-sampling process on the latent
projection path. Second, it outputs the mask embedding that informs
the latent projection layer the feature clusters that are most likely
coherent with the particular mask. To reflect the fact that mask
features from the left contracting path serves mainly as a constraint,
only 8 mask features are concatenated to the latent projection path of the
network. This design is based on two reasons: (1) more mask features require more capacity to properly merge them into
projected latent features, increasing training difficulty; (2)
our preliminary experiments indicate a trained model mostly project
mask features with almost identical patterns merely in different
numerical values.


\section{Training}

All three models are trained using the WGAN-GP loss
formulation\cite{wgan_gp}. The Pix2Pix baseline was trained directly
at target resolution for 25 epochs. Our proposed model with and
without mask embedding was trained using the progressive growing
strategy from the pGAN study \cite{karras2018progressive}. We start
from an output resolution of $8^{2}$, train for 45k steps and then
fade in new convolution blocks that doubles the input and
output resolution. Given the light-weightiness of the mask projection
path, no fading connection is implemented.



To compare the effectiveness of mask embedding mechanism, the
training schedule including batch size, learning rate and number of
discriminator optimization per generator in this study is kept the
same for our proposed models. We used a batch size of 256 at the
output resolution of $8\times8$. We half the batch size every time
when doubling the output resolution. The learning rate is initially
set to constant at $0.001$ and increase to $0.002$ when the model
reaches the output resolution of $256$. More details can be found
in our source code. We use TensorFlow platform and each model in our
experiment is trained on 4 NVIDIA V100 for 2 days to reach the final
resolution of $512^2$.

\section{Experiments}

We compared the generators of \textbf{Pix2Pix baseline} ($23.23M$), our \textbf{without embedding baseline} ($23.07M$), and our \textbf{proposed embedding model} ($23.79M$) on an image synthesis task using the CELEBA-HQ dataset. The pix2pix-HD model is not compared due to the fact that its instance-level feature embedding mechanism depends on the perturbation of masks to generate diverse images, while our proposed method focus on one to more mapping of the same semantic mask. Hyper-parameters of compared models are intentionally kept very similar at the latent projection path (up-sampling path for Pix2Pix) for controlled performance comparison. The discriminators of both our baseline and proposed model are identical
containing $23.07M$ parameters.Generator performance are measured using sliced Wasserstein distance(SWD).


\subsection{CELEBA-HQ Dataset}

The dataset we used to validate our approach is the CELEBA-HQ dataset
originally compiled by \cite{celeba},  later cleanup and augmented
by \cite{karras2018progressive}. We extracted 68 face landmarks
for each face images in CELEBA-HQ dataset using the Dlib Python
Library\cite{dlib09}. The detection is performed at resolution of
$1024^2$. We then constructed the face edge map simply by connecting
the detected dots from each face landmark. Landmark detections
significantly different from the original specified attributes
\cite{celeba} were removed. In total 27000 images were compiled as
training images.

\subsection{Quantitative Evaluation} 

We evaluated the effectiveness of proposed model using the sliced
Wasserstein distance SWD \cite{SWD}, following the parameter settings
used previously \cite{karras2018progressive}. Due to memory limitation,
SWD was averaged over batches of real and synthesized images pairs. We
first computed the SWD of 240 image pairs and then repeated until we
cover 8192 pairs. We generated the Laplacian pyramid of the images
from $512^2$ to the minimal resolution of $16^2$. At each level of
the pyramid we extracted 128 patches of size $7\times7$, normalized
and computed the average distance for each level with respect the
real images. In Table~\ref{SWD_table}, the SWD metric captures the
performance difference of our baseline and proposed models, as well
as the Pix2Pix model. We can infer that using masking embedding is
superior and improves the quality of synthesized images, which is
also consistent with the visual observations.

\begin{table}
  \centering
  \begin{tabular}{llllllll}
    \toprule
    Configurations     &   512 &   256 &   128 &    64 &    32 &    16 & Avg \\
    \midrule
    Real              & 10.82 &  9.98 & 10.14 &  9.75 &  9.83 &  7.52 &  9.67 \\
    Pix2Pix           &67.74  &  27.72 &  25.08   &  20.46  & 19.05 &  151.78 & 65.52    \\
    Without Embedding & 58.20 & 27.77 & 22.19 & 18.25 & 17.58 & 70.49 & \textbf{35.75}  \\
    With Embedding    & 43.74 & 22.46 & 17.48 & 14.83 & 13.65 & 37.57 & \textbf{24.96}  \\
    \bottomrule
  \end{tabular}
  \caption{Sliced Wasserstein distance(SWD) measured between the
  generated images of our baseline and proposed model to the training
  images. Each column is one level on the Laplacian pyramid.}
  \label{SWD_table}
\end{table}

\subsection{Qualitative Comparison}

\setlength\tabcolsep{0.25pt}
\begin{figure}[!ht]
\centering
\begin{tabular}{ccccc}
     \includegraphics[width=0.2\textwidth]{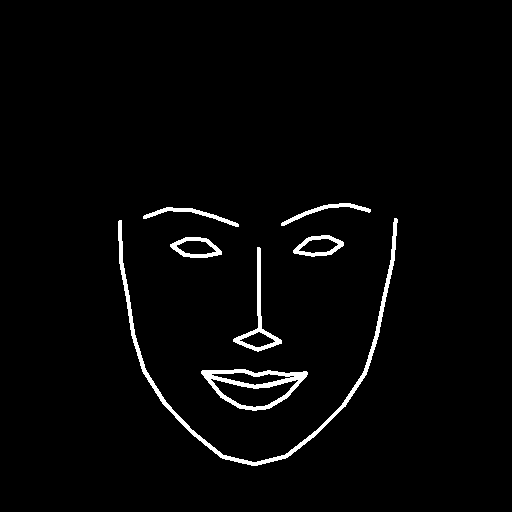}&
     \includegraphics[width=0.2\textwidth]{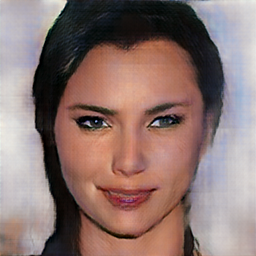}&
     \includegraphics[width=0.2\textwidth]{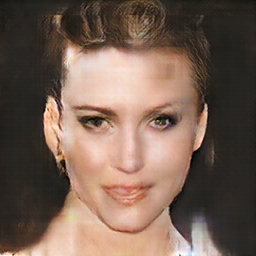}&
     \includegraphics[width=0.2\textwidth]{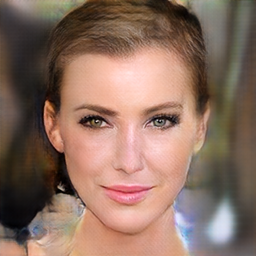}&\\
     \includegraphics[width=0.2\textwidth]{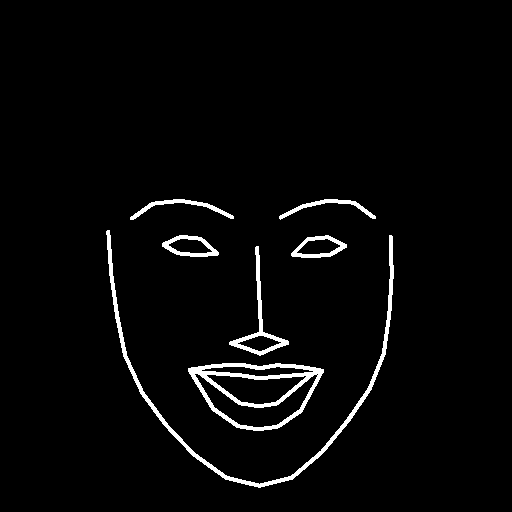}&  
     \includegraphics[width=0.2\textwidth]{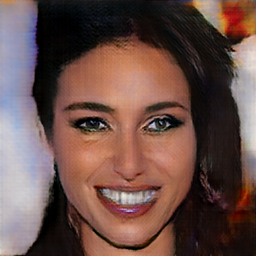}&
     \includegraphics[width=0.2\textwidth]{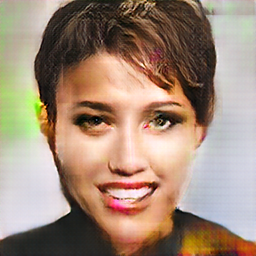}&
     \includegraphics[width=0.2\textwidth]{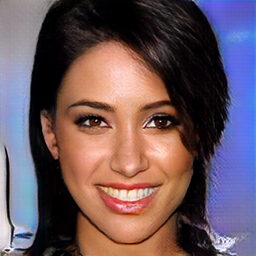}\\     \includegraphics[width=0.2\textwidth]{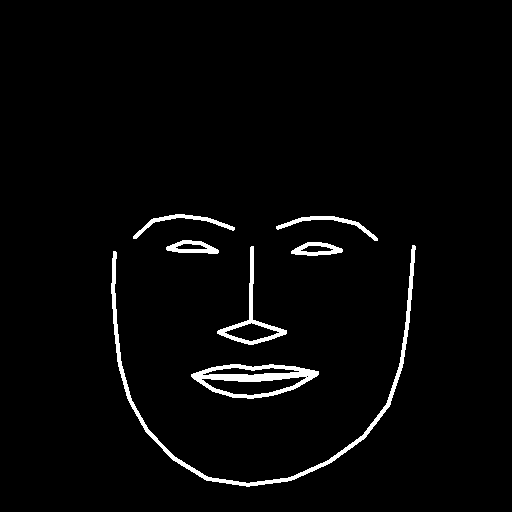}&  
     \includegraphics[width=0.2\textwidth]{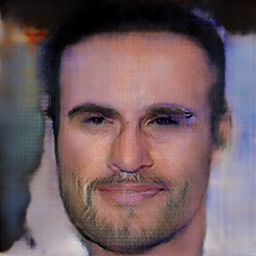}&
     \includegraphics[width=0.2\textwidth]{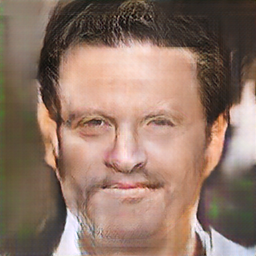}&
     \includegraphics[width=0.2\textwidth]{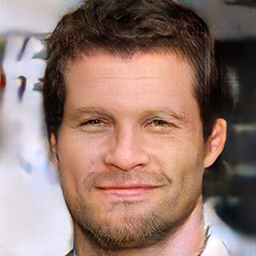}\\     \includegraphics[width=0.2\textwidth]{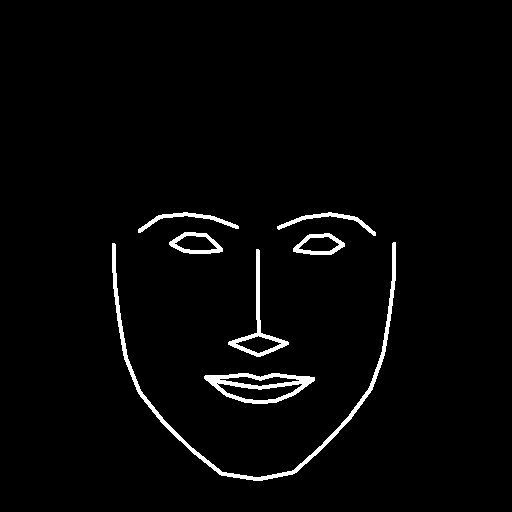}&            \includegraphics[width=0.2\textwidth]{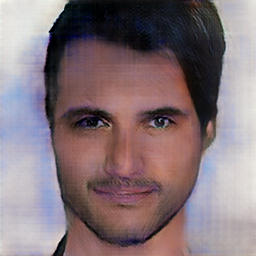}&
     \includegraphics[width=0.2\textwidth]{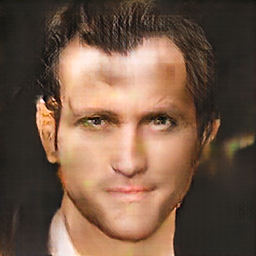}&
     \includegraphics[width=0.2\textwidth]{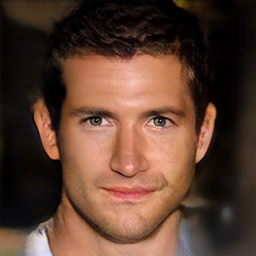}\\
     (a)&(b)&(c)&(d)
\end{tabular}
\caption{(a) Input mask. (b) Synthesized image using Pix2Pix (c)
Synthesized image using our without-embedding baseline model. (d)
Synthesized image using our proposed embedding model.}
\label{fig:Embedding_No_Embedding_Comparsion}
\end{figure}

Fig~\ref{fig:Embedding_No_Embedding_Comparsion} illustrates the
synthesized results of the two baseline models and our proposed embedding model
using the same mask as input. Compared to our proposed model, the
Pix2Pix baseline is limited to generate coarse images in very similar style. For example, a particular model iteration during training generates only black or dark brown hair color, or the skin texture has the same waxy appearance. In this baseline Pix2Pix model, it is likely that the color and texture of faces are strongly coupled with the mask input, forcing the model to learn only the 'average' face in the dataset, thus
preventing the model from synthesizing diverse, fine-grained textures.

The model without embedding also failed to generate high fidelity
textures. The generated images contain major noise realization
and up-sampling artifacts that are indications of reduction in model
capacity. This observation fits our hypothesis that the model without
mask embedding is forced to project initial features onto space at the intersection
of sample distributions, resulting in blurred texture patterns and
ambiguous structures. As a consequence of insufficient generator
capacity during training, the model also generates significantly
more artifacts such as diagonal straight lines and checkerboard
texture patterns.

\subsection{Changing Latent Input}

\begin{figure}[!ht]
\begin{tabular}{ccccc}
     \includegraphics[width=0.2\textwidth]{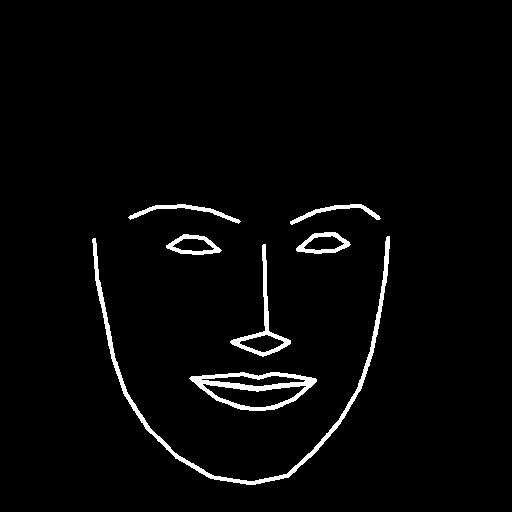}&  \includegraphics[width=0.2\textwidth]{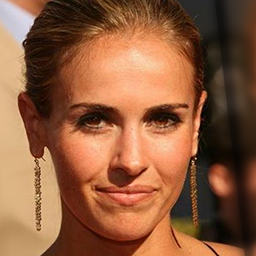}& \includegraphics[width=0.2\textwidth]{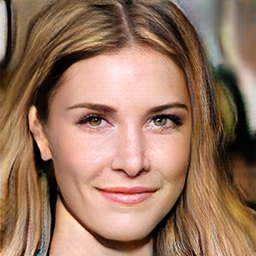}& \includegraphics[width=0.2\textwidth]{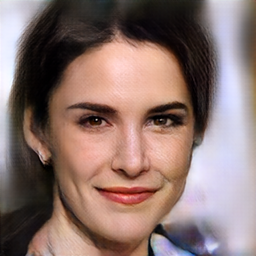}& \includegraphics[width=0.2\textwidth]{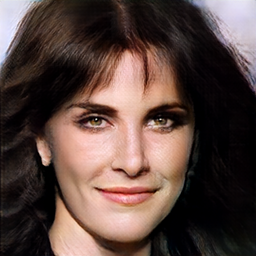} \\
     
     \includegraphics[width=0.2\textwidth]{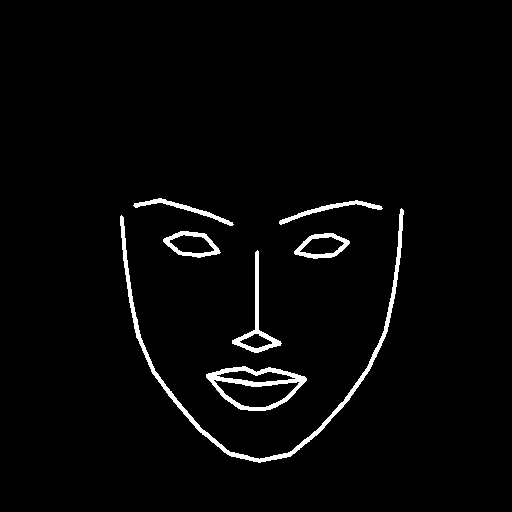}&  \includegraphics[width=0.2\textwidth]{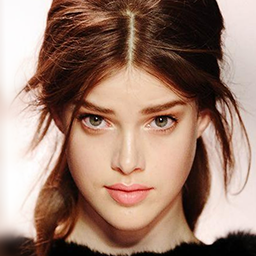}& \includegraphics[width=0.2\textwidth]{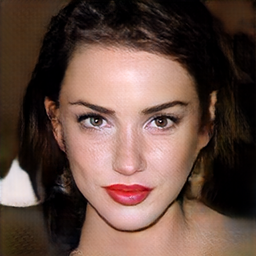}& \includegraphics[width=0.2\textwidth]{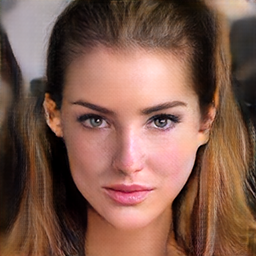}& \includegraphics[width=0.2\textwidth]{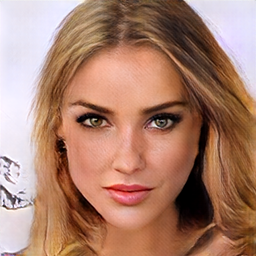} \\
     
     \includegraphics[width=0.2\textwidth]{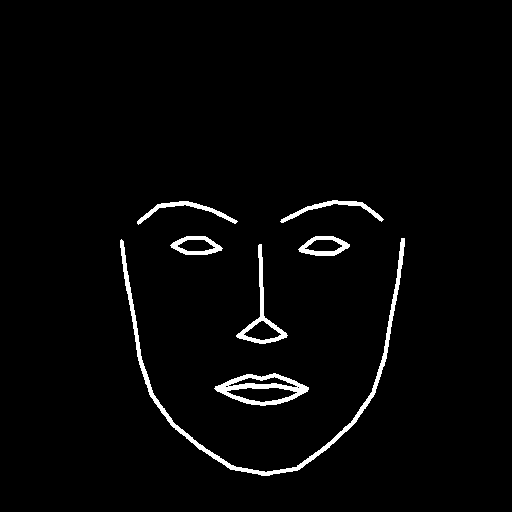}&  \includegraphics[width=0.2\textwidth]{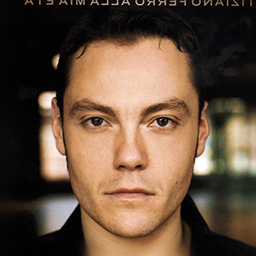}& \includegraphics[width=0.2\textwidth]{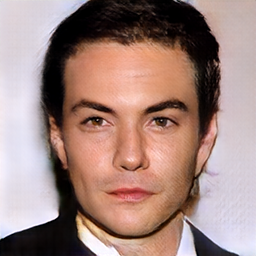}& \includegraphics[width=0.2\textwidth]{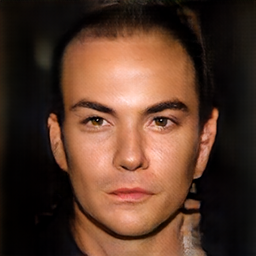}& \includegraphics[width=0.2\textwidth]{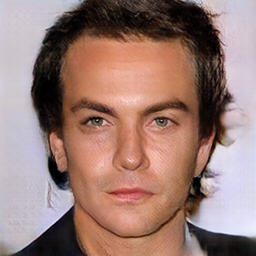} \\
     
     \includegraphics[width=0.2\textwidth]{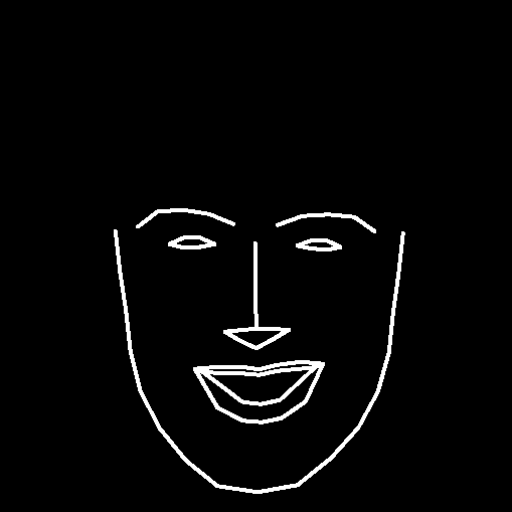}&  \includegraphics[width=0.2\textwidth]{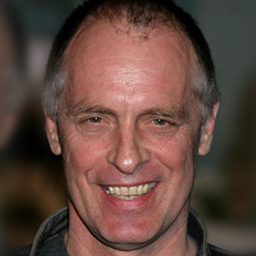}& \includegraphics[width=0.2\textwidth]{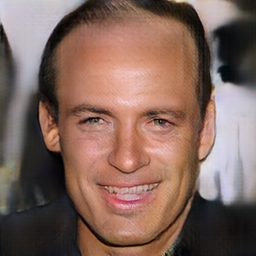}& \includegraphics[width=0.2\textwidth]{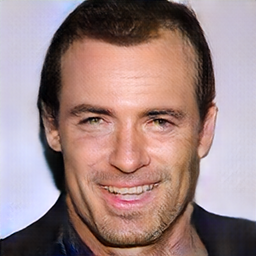}& \includegraphics[width=0.2\textwidth]{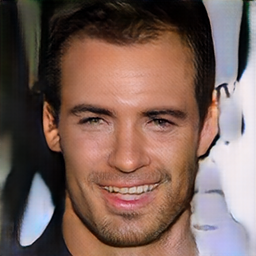} \\
     (a)&(b)&(c)&(d)&(e)
\end{tabular}
\caption{(a) Input mask. (b) Original Image. (c), (d), (e) synthesized images using the same mask but different latent vectors}
\label{fig:Changing_Latent_Input}
\end{figure}

We also demonstrate that the same mask input can be coupled with different latent vectors to form different faces in Fig
\ref{fig:Changing_Latent_Input}. We noticed however that the latent
vector and mask embedding were not completely disentangled. The
latent vector is responsible more for the style of images, namely
the hair style, skin color, facial hair, etc. On the other hand, the
face landmarks are as expected determined by the provided mask. More results can be found in supplemental material. One
limitation of this study is the small number of images compared to
varieties of facial landmarks combinations generated. We observe some
masks are coupled with specific characteristics such as gender and skin
color that not necessarily obvious to human observer given a binary
mask. This prevented the latent vector gaining control for better
sample space mapping. For future work, this problem could potentially
be alleviated using more abstract mask input together with a larger
dataset. Moreover, implementing random blurring and mask feature drop
output could potential help increase the output variety as well. 

\section{Conclusion}

We have demonstrated the significance of mask embedding in
high-resolution realistic image synthesis. Quantitative and qualitative
evaluations validate the effectiveness of mask embedding mechanism. Our
experiment is based on semantic input, and the same concept applies
to other conditional input such as textures and text.

\clearpage
\bibliographystyle{unsrt}
\bibliography{reference}

\begin{thebibliography}{10}

\bibitem{Pix2pix_HD}
Ting-Chun Wang, Ming-Yu Liu, Jun-Yan Zhu, Andrew Tao, Jan Kautz, and Bryan
  Catanzaro.
\newblock High-resolution image synthesis and semantic manipulation with
  conditional gans.
\newblock In {\em Proceedings of the IEEE Conference on Computer Vision and
  Pattern Recognition}, 2018.

\bibitem{refinementGAN}
Qifeng Chen and Vladlen Koltun.
\newblock Photographic image synthesis with cascaded refinement networks.
\newblock In {\em {IEEE} International Conference on Computer Vision, {ICCV}
  2017, Venice, Italy, October 22-29, 2017}, pages 1520--1529, 2017.

\bibitem{stackGAN}
Han Zhang, Tao Xu, Hongsheng Li, Shaoting Zhang, Xiaogang Wang, Xiaolei Huang,
  and Dimitris Metaxas.
\newblock Stackgan: Text to photo-realistic image synthesis with stacked
  generative adversarial networks.
\newblock In {\em {ICCV}}, 2017.

\bibitem{wGAN}
Martin Arjovsky, Soumith Chintala, and L{\'e}on Bottou.
\newblock {W}asserstein generative adversarial networks.
\newblock In Doina Precup and Yee~Whye Teh, editors, {\em Proceedings of the
  34th International Conference on Machine Learning}, volume~70 of {\em
  Proceedings of Machine Learning Research}, pages 214--223, International
  Convention Centre, Sydney, Australia, 06--11 Aug 2017. PMLR.

\bibitem{wgan_gp}
Ishaan Gulrajani, Faruk Ahmed, Martin Arjovsky, Vincent Dumoulin, and Aaron~C
  Courville.
\newblock Improved training of wasserstein gans.
\newblock In I.~Guyon, U.~V. Luxburg, S.~Bengio, H.~Wallach, R.~Fergus,
  S.~Vishwanathan, and R.~Garnett, editors, {\em Advances in Neural Information
  Processing Systems 30}, pages 5767--5777. Curran Associates, Inc., 2017.

\bibitem{pix2pix2016}
Phillip Isola, Jun-Yan Zhu, Tinghui Zhou, and Alexei~A Efros.
\newblock Image-to-image translation with conditional adversarial networks.
\newblock {\em arxiv}, 2016.

\bibitem{UNet}
Olaf Ronneberger, Philipp Fischer, and Thomas Brox.
\newblock U-net: Convolutional networks for biomedical image segmentation.
\newblock In {\em MICCAI}, 2015.

\bibitem{Tube_GAN}
He~Zhao, Huiqi Li, Sebastian Maurer-Stroh, and Li~Cheng.
\newblock Synthesizing retinal and neuronal images with generative adversarial
  nets.
\newblock {\em Medical Image Analysis}, 49:14 -- 26, 2018.

\bibitem{ConditionalGA}
Mehdi Mirza and Simon Osindero.
\newblock Conditional generative adversarial nets.
\newblock {\em CoRR}, abs/1411.1784, 2014.

\bibitem{cGAN_face_attribute}
Jon Gauthier.
\newblock Conditional generative adversarial nets for convolutional face
  generation.
\newblock 2015.

\bibitem{cGAN_facial_expression}
Xueping Wang, Weixin Li, Guodong Mu, Di~Huang, and Yunhong Wang.
\newblock Facial expression synthesis by u-net conditional generative
  adversarial networks.
\newblock In {\em Proceedings of the 2018 ACM on International Conference on
  Multimedia Retrieval}, ICMR '18, pages 283--290, New York, NY, USA, 2018.
  ACM.

\bibitem{cGAN_face_aging}
G.~{Antipov}, M.~{Baccouche}, and J.~{Dugelay}.
\newblock Face aging with conditional generative adversarial networks.
\newblock In {\em 2017 IEEE International Conference on Image Processing
  (ICIP)}, pages 2089--2093, Sep. 2017.

\bibitem{cGAN_histology}
N.~{Bayramoglu}, M.~{Kaakinen}, L.~{Eklund}, and J.~{Heikkilä}.
\newblock Towards virtual h e staining of hyperspectral lung histology images
  using conditional generative adversarial networks.
\newblock In {\em 2017 IEEE International Conference on Computer Vision
  Workshops (ICCVW)}, pages 64--71, Oct 2017.

\bibitem{cGAN_brain_tumor}
Mina Rezaei, Konstantin Harmuth, Willi Gierke, Thomas Kellermeier, Martin
  Fischer, Haojin Yang, and Christoph Meinel.
\newblock A conditional adversarial network for semantic segmentation of brain
  tumor.
\newblock In Alessandro Crimi, Spyridon Bakas, Hugo Kuijf, Bjoern Menze, and
  Mauricio Reyes, editors, {\em Brainlesion: Glioma, Multiple Sclerosis, Stroke
  and Traumatic Brain Injuries}, pages 241--252, Cham, 2018. Springer
  International Publishing.

\bibitem{cGAN_MRI}
S.~U. {Dar}, M.~{Yurt}, L.~{Karacan}, A.~{Erdem}, E.~{Erdem}, and T.~{Çukur}.
\newblock Image synthesis in multi-contrast mri with conditional generative
  adversarial networks.
\newblock {\em IEEE Transactions on Medical Imaging}, pages 1--1, 2019.

\bibitem{text_embedding}
X.~{Liu}, G.~{Meng}, S.~{Xiang}, and C.~{Pan}.
\newblock Semantic image synthesis via conditional cycle-generative adversarial
  networks.
\newblock In {\em 2018 24th International Conference on Pattern Recognition
  (ICPR)}, pages 988--993, Aug 2018.

\bibitem{disentanglement}
G{\"o}khan {Yildirim}, Calvin {Seward}, and Urs {Bergmann}.
\newblock {Disentangling Multiple Conditional Inputs in GANs}.
\newblock {\em arXiv e-prints}, page arXiv:1806.07819, Jun 2018.

\bibitem{Thermal_pix2pix}
Cunjian {Chen} and Arun {Ross}.
\newblock {Matching Thermal to Visible Face Images Using a Semantic-Guided
  Generative Adversarial Network}.
\newblock {\em arXiv e-prints}, page arXiv:1903.00963, Mar 2019.

\bibitem{Retinal_GAN}
P.~{Costa}, A.~{Galdran}, M.~I. {Meyer}, M.~{Niemeijer}, M.~{Abràmoff}, A.~M.
  {Mendonça}, and A.~{Campilho}.
\newblock End-to-end adversarial retinal image synthesis.
\newblock {\em IEEE Transactions on Medical Imaging}, 37(3):781--791, March
  2018.

\bibitem{karras2018progressive}
Tero Karras, Timo Aila, Samuli Laine, and Jaakko Lehtinen.
\newblock Progressive growing of {GAN}s for improved quality, stability, and
  variation.
\newblock In {\em International Conference on Learning Representations}, 2018.

\bibitem{DCGAN}
Joachim~D. Curtó, Irene~C. Zarza, Fernando De~La~Torre, Irwin King, and
  Michael~R. Lyu.
\newblock High-resolution deep convolutional generative adversarial networks,
  2017.
\newblock cite arxiv:1711.06491.

\bibitem{style_gan}
Tamar {Rott Shaham}, Tali {Dekel}, and Tomer {Michaeli}.
\newblock {SinGAN: Learning a Generative Model from a Single Natural Image}.
\newblock {\em arXiv e-prints}, page arXiv:1905.01164, May 2019.

\bibitem{pGAN_Brain}
Anthreas Antoniou, Amos Storkey, and Harrison Edwards.
\newblock Data augmentation generative adversarial networks, 2018.

\bibitem{pgan_mammo}
Dimitrios {Korkinof}, Tobias {Rijken}, Michael {O'Neill}, Joseph {Yearsley},
  Hugh {Harvey}, and Ben {Glocker}.
\newblock {High-Resolution Mammogram Synthesis using Progressive Generative
  Adversarial Networks}.
\newblock {\em arXiv e-prints}, page arXiv:1807.03401, Jul 2018.

\bibitem{celeba}
Ziwei Liu, Ping Luo, Xiaogang Wang, and Xiaoou Tang.
\newblock Deep learning face attributes in the wild.
\newblock In {\em Proceedings of International Conference on Computer Vision
  (ICCV)}, December 2015.

\bibitem{dlib09}
Davis~E. King.
\newblock Dlib-ml: A machine learning toolkit.
\newblock {\em Journal of Machine Learning Research}, 10:1755--1758, 2009.

\bibitem{SWD}
Rabin Julien, Gabriel Peyr{\'e}, Julie Delon, and Bernot Marc.
\newblock {Wasserstein Barycenter and its Application to Texture Mixing}.
\newblock In {\em {SSVM'11}}, pages 435--446, Israel, 2011. {Springer}.

\end{thebibliography}

\end{document}